\documentclass{article}

\usepackage{arxiv}
\usepackage[utf8]{inputenc}
\usepackage[T1]{fontenc}
\usepackage{hyperref}
\usepackage{url}
\usepackage{booktabs}
\usepackage{amsfonts}
\usepackage{nicefrac}
\usepackage{microtype}
\usepackage{graphicx}

\usepackage{doi}

\title{Beyond RNNs: Benchmarking Attention-Based Image Captioning Models}

\author{
    Hemanth Teja Yanambakkam \\
    New York University \\
    \texttt{hy1713@nyu.edu} \\
    \And
    Rahul Chinthala \\
    New York University
}

\begin{document}
\maketitle
\thispagestyle{empty}

\section{Problem Statement}
 
Image captioning is the automatic generation of textual descriptions for images, a fundamental challenge in artificial intelligence at the intersection of computer vision and natural language processing. This task involves identifying key objects, their attributes, and their relationships within an image while generating syntactically and semantically coherent sentences that accurately describe the scene. Deep learning-based techniques, known for their effectiveness in various complex tasks, have also been applied to image captioning.

In this study, we conduct a comparative analysis of two deep learning-based image captioning models using the MS-COCO benchmark dataset. Our goal is to evaluate their performance using natural language processing metrics.

\section{Literature Survey}

In this section, we provide relevant background on previous work in image caption generation and attention mechanisms. Extensive research has been conducted on automatic image captioning, which can broadly be categorized into three approaches: template-based captioning, retrieval-based captioning, and novel image caption generation \cite{xu2015}.
  
Template-based image captioning first detects objects, attributes, and actions in an image before filling predefined slots in a fixed template \cite{shetty2017}. Retrieval-based approaches, on the other hand, identify visually similar images from the training dataset and select a caption from the retrieved images \cite{lin2014}. While these methods can generate syntactically correct captions, they often fail to produce semantically meaningful and image-specific descriptions. Both approaches typically involve an intermediate “generalization” step to remove details that are only relevant to a specific retrieved image, such as the name of a city. However, these methods have largely been replaced by neural network-based approaches, which have become the dominant paradigm in image captioning.

In contrast, novel approaches to image caption generation first analyze the visual content of an image before generating captions using a language model \cite{kuznetsova2012}. The visual representation is typically extracted using a convolutional neural network (CNN), which is often pre-trained on large-scale image classification datasets \cite{karpathy2017}. The caption is then generated using a recurrent neural network (RNN)-based language model. The key advantage of this approach is that the entire system can be trained end-to-end, allowing all parameters to be learned directly from the data.

The first approach to using neural networks for caption generation was proposed by Kiros et al. (2014a), who introduced a multimodal log-bilinear model biased by image features. This work was later extended by Kiros et al. (2014b). Mao et al. (2014) adopted a similar approach but replaced the feedforward neural language model with a recurrent one. Both Vinyals et al. (2014) and Donahue et al. (2014) employed recurrent neural networks (RNNs) based on long short-term memory (LSTM) units (Hochreiter \& Schmidhuber, 1997) for their models.

The problem with these methods is that when the model attempts to generate the next word in the caption, it usually describes only a specific part of the image. As a result, it fails to capture the essence of the entire input image. Conditioning the generation of each word on the whole image representation does not effectively produce distinct words for different parts of the image. This is precisely where an attention mechanism proves useful.

\section{Description of Dataset}

The MS-COCO (Common Objects in Context) dataset is used in this project. It is widely utilized for training and benchmarking object detection, segmentation, and captioning algorithms. Some key features of the dataset include 80 object categories, 91 stuff categories, 330K images, and five captions per image. Since the dataset has fewer categories but a higher number of instances per category, we have chosen it for training our model.

\section{Data Preprocessing}
We tokenized the captions and performed basic cleaning, such as lowercasing all words and removing special tokens. Furthermore, we created a vocabulary of all unique words present across the image captions, resulting in a total of 40,000 data points (8,000 × 5). After filtering, 8,267 unique words remained.

We then selected the top 5,000 or 7,000 words from the vocabulary to improve the model’s robustness to outliers and reduce errors. According to PyTorch and Keras documentation, pre-trained Inception V3 and ResNet50 models require input as 3-channel RGB images of shape (3 × H × W), where H and W must be at least 224. Therefore, data preprocessing involves loading and resizing images. The tokens and images are then converted into embeddings of size 256, which are fed to the decoder.

\section{Description of Models Implemented}

\subsection{Vanilla RNN Image Captioner}

Here, we use CNN and RNN to generate image captions, with CNN acting as the encoder and RNN as the decoder.

\subsubsection{Encoder-CNN}

To generate a description, we feed an image into a pre-trained CNN architecture, often called the encoder, as it encodes the image content into a smaller feature vector. A CNN scans images from left to right and top to bottom, extracting important features and combining them to classify images.

At the end of the CNN network is a softmax classifier that outputs a vector of class scores. However, instead of classifying the image, we aim to extract a set of features that represent its spatial content. To achieve this, we remove the final fully connected layer responsible for classification and instead focus on an earlier layer that captures the spatial information.

This approach allows us to use the CNN as a feature extractor, compressing the vast amount of information in the original image into a more compact representation.

\begin{figure}[htp]
    \centering
    \includegraphics[width=8cm]{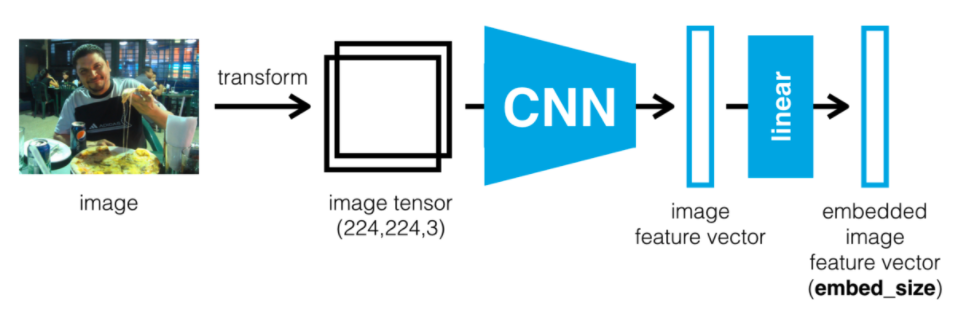}
    \caption{Architecture of the CNN Encoder}
    \label{fig:cnn_encoder}
\end{figure}

\subsubsection{Decoder-RNN}

The job of the RNN is to decode the processed vector output by the CNN and transform it into a sequence of words. Consequently, this part of the network is often referred to as the decoder.

\begin{figure}[htp]
    \centering
    \includegraphics[width=8cm]{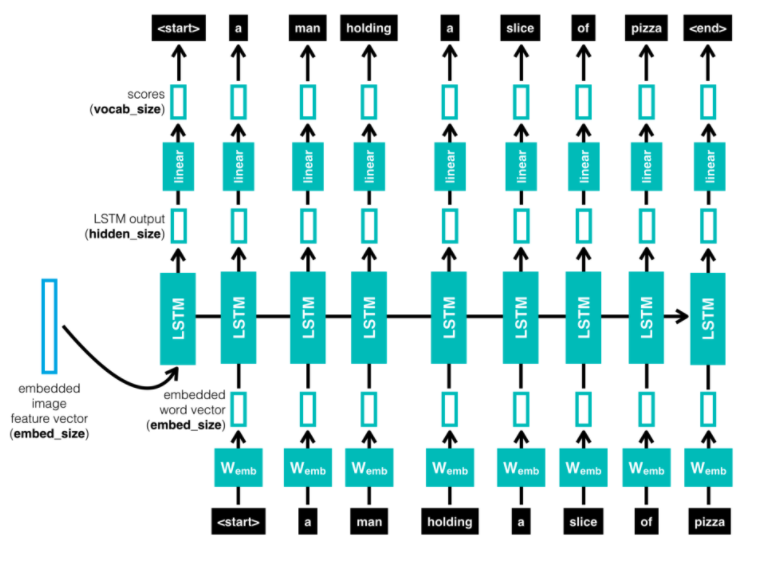}
    \caption{Architecture of the CNN Decoder}
    \label{fig:cnn_decoder}
\end{figure}

\subsection{Our Model's Three Main Phases}

\begin{itemize}
    \item \textbf{Image Feature Extraction:} The features of images from the MS COCO dataset are extracted using the ResNet50 model. We obtain a 2048-element vector representation of each photo and pass it on to the LSTM layer.
    
    \item \textbf{Sequence Processor:} The sequence processor handles text input by acting as a word embedding layer. This embedding layer applies rules to extract the necessary text features and includes a mask to ignore padded values. The network is then connected to an LSTM for the final phase of image captioning.
    
    \item \textbf{Decoder:} The final phase of the model combines the input from the image extractor and sequence processor phases using an additional operation. The result is then fed into a 256-neuron layer, followed by a final output dense layer that generates a softmax prediction of the next word in the caption. This prediction is made over the entire vocabulary, which was constructed from the text data processed in the sequence processor phase.
\end{itemize}

\subsection{Loss Function}

The nature of our RNN output is a sequence of likelihoods representing word occurrences. To quantify the quality of the RNN output, we use Cross-Entropy Loss, which is the most widely used and effective metric for evaluating the performance of a classification model whose output is a probability value between 0 and 1.

\begin{equation}
Loss = - \sum_{n=1}^{N} t^n_k \ln  p^n_k
\end{equation}

Here, \(N\) is the number of classes, representing the vocabulary size in this context, \(t\) is either 0 or 1, and \(p^{n}_{k}\) is the predicted probability that observation \(k\) belongs to class \(n\).

\section{Attention-Based Image Captioning}

In this technique, the image is first divided into \( n \) parts, and we compute Convolutional Neural Network (CNN) representations for each part, \( h_1, \dots, h_n \). When the RNN generates a new word, the attention mechanism allows it to focus on the part of the image most relevant to the word it is about to generate. The model learns where to look, and as we generate a caption word by word, we can observe the model’s focus shifting across different regions of the image.

\begin{equation}
\sum_{t=1}^{T} \alpha_{p,t} \approx 1
\end{equation}

\subsection{Bahdanau Attention}

In this project, we have used Bahdanau Attention, also known as Additive Attention, as it performs a linear combination of the encoder states and the decoder states. In the Bahdanau Attention mechanism, all encoder hidden states, along with the decoder hidden state, are used to generate the context vector.

We calculate the alignment scores between the previous decoder hidden state and each of the encoder’s hidden states. These alignment scores for each encoder hidden state are combined into a single vector and then passed through a softmax function. The alignment vector has the same length as the source sequence, with each value representing the score (or probability) of the corresponding word in the source sequence.

Alignment vectors assign weights to the encoder’s output, allowing the decoder to determine which parts of the input to focus on at each time step. The encoder hidden states and their respective alignment scores (attention weights in the above equation) are multiplied to form the context vector, which is then used to compute the final output of the decoder.

\subsection{Training Details}

\textbf{ResNet-50:} It is a convolutional neural network that is 50 layers deep, consisting of 48 convolutional layers, along with one MaxPool and one Average Pool layer. It performs approximately \(3.8 \times 10^9\) floating-point operations. 

\begin{itemize}
    \item[] \textbf{ResNet-50:} It is a convolutional neural network that is 50 layers deep, consisting of 48 convolutional layers, along with one MaxPool and one Average Pool layer. It performs approximately \(3.8 \times 10^9\) floating-point operations.
    
    \item[] \textbf{Inception V3:} Inception v3 is a convolutional neural network that is 48 layers deep and designed to reduce computational cost by optimizing previous Inception architectures.

    \item[] \textbf{Teacher Forcing:} We have used teacher forcing instead of feeding the output of a time step into the next time step, as it allows recurrent neural network models to train more quickly and efficiently by using the ground truth from a prior time step as input.

    \item[] \textbf{Vocab Size:} We have varied the vocabulary size (Top 5000, Top 7000), as ideally, we want a vocabulary that is both expressive and as small as possible. A smaller vocabulary results in a more compact model that trains faster.

    \item[] \textbf{Learning Rate:} We adjusted several hyperparameters during our experiments. Initially, in the training phase, we set the learning rate to \(4 \times 10^{-4}\). However, under this condition, the training loss exhibited significant oscillations.

    \item[] \textbf{Batch Size:} We also adjusted the batch size in our experiments. Initially, we used a batch size of 16. However, since the loss did not converge, we increased the batch size to 64, as we were constrained by the 12GB maximum GPU limit offered by Colab.

    \item[] \textbf{Weight Initialization:} We have used Glorot/Xavier initialization for the weights as it helps maintain similar gradients, Z-values, and activations across all layers. Additionally, the Sigmoid activation function presents a challenge in this context, as its activation values have a mean of 0.5 rather than zero.
\end{itemize}

\section{Evaluation Metrics}
BLEU-1, BLEU-2, BLEU-3, and BLEU-4 (Bilingual Evaluation Understudy) are among the most commonly reported metrics for evaluating the quality of text generation in natural language processing tasks. Here, we chose the 4-gram BLEU score (BLEU-4) as our primary evaluation metric. Furthermore, we also use the METEOR score, which is based on the harmonic mean of unigram precision and recall, with recall weighted higher than precision. 

Additionally, we use GLEU (for sentence-level fluency) and WER (Word Error Rate), which are common metrics for evaluating the performance of speech recognition and machine translation systems.
\section{Results}

\subsection{Model: Bahdanau Attention + Inception V3}

It can be observed that the METEOR score increases as the vocabulary size and the number of epochs increase, meaning that the model performs better when trained with higher configuration parameters and more data. 

Moreover, the configuration with:
\begin{itemize}
    \item Vocabulary size: 7000
    \item Number of epochs: 30
    \item Number of images: 6000
\end{itemize}
achieved the best BLEU-1,2,3,4 and GLEU scores. 

However, the scores decreased for the configuration with:
\begin{itemize}
    \item Vocabulary size: 7000
    \item Number of epochs: 50
    \item Number of images: 10,000
\end{itemize}
This suggests possible overfitting of the model.

\begin{figure}[htbp]
    \centering
    \includegraphics[width=0.92\textwidth]{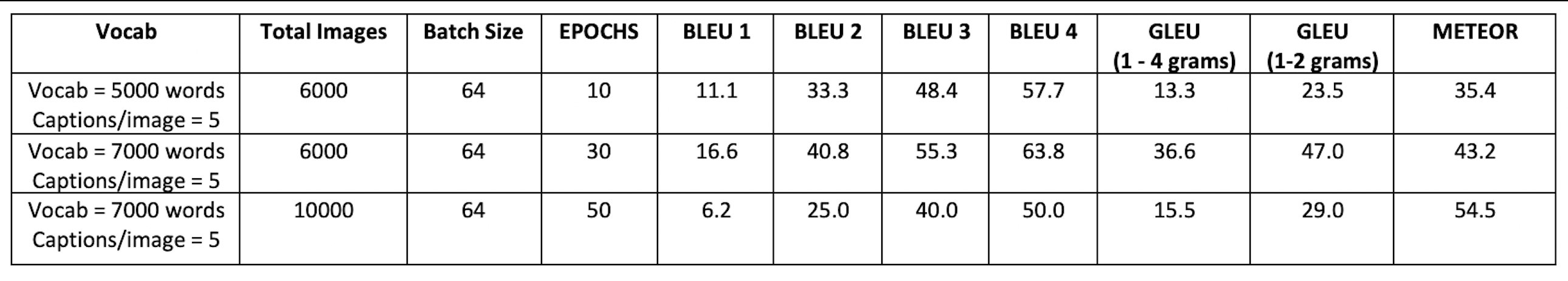}
    \caption{Results for Model 1 - Bahdanau Attention + Inception V3}
    \label{fig:model1_res1}
\end{figure}

\begin{figure}[htbp]
    \centering
    \includegraphics[width=0.92\textwidth]{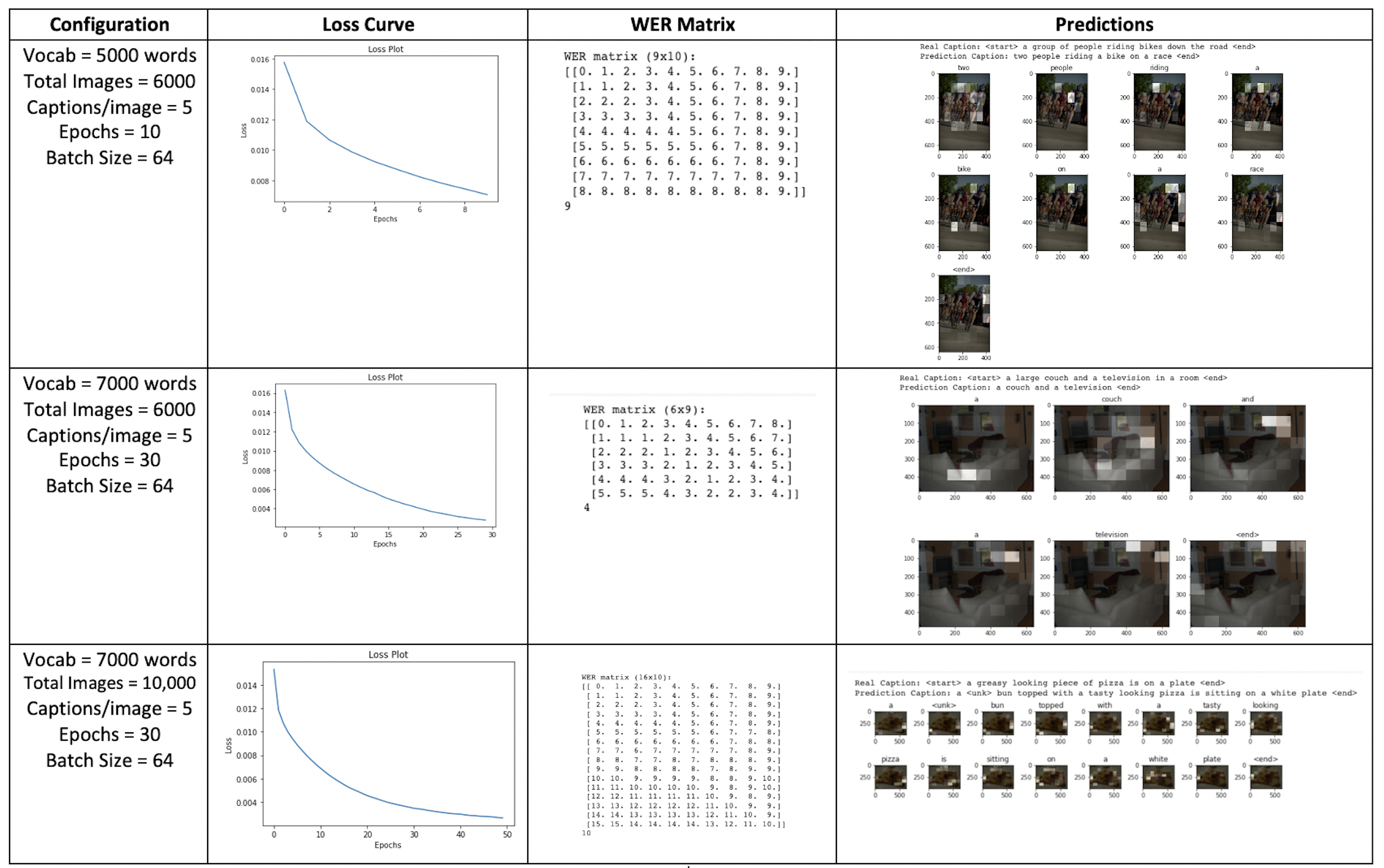}
    \caption{Results for Model 1 - Bahdanau Attention + Inception V3 (Continued)}
    \label{fig:model1_res2}
\end{figure}

\subsection{Model: Bahdanau Attention + ResNet50}

We have inferred that, in some cases, METEOR scores do not align well with human judgments. For example, while human evaluators rated a particular image as “Good,” its METEOR score was lower (30.9) than another image that was rated as “Fair” (37.5). This discrepancy suggests that METEOR, which primarily emphasizes word similarity, may not always capture the full semantic accuracy and coherence of an image caption.

A possible explanation for this discrepancy is that:
\begin{itemize}
    \item The model captured many nuances in the image and the corresponding words.
    \item However, it neglected the semantic order of the words.
\end{itemize}
Since METEOR primarily emphasizes word similarity, this resulted in:
\begin{itemize}
    \item A high METEOR score
    \item A lower human judgment score due to lack of semantic coherence
\end{itemize}

\begin{figure}[htbp]
    \centering
    \includegraphics[width=0.99\textwidth]{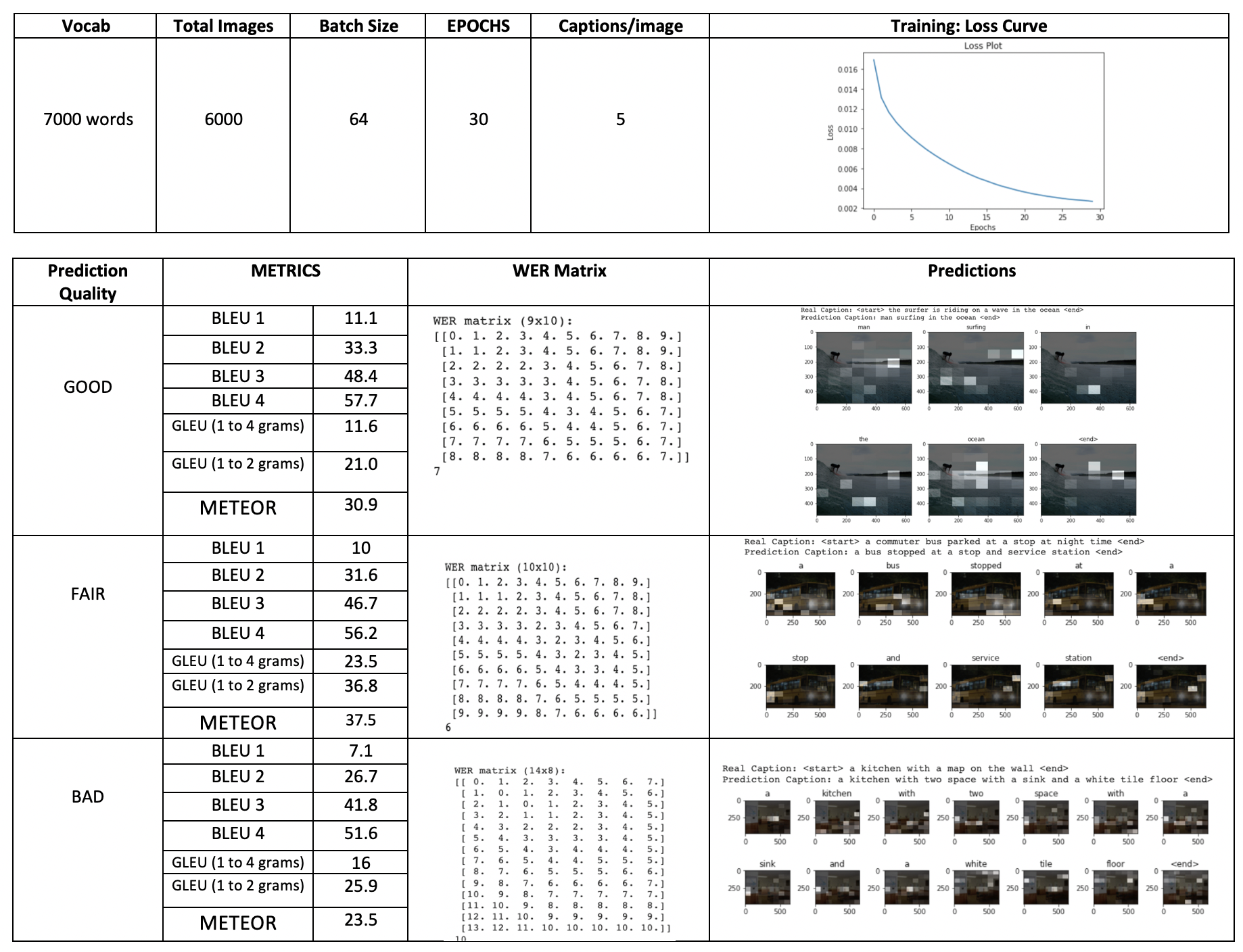}
    \caption{Results for Model 2 - Bahdanau Attention + ResNet50}
    \label{fig:model2_res}
\end{figure}

\noindent\textbf{View:} \href{https://github.com/HemanthTejaY/Image-Captioning-A-Comparative-Study/blob/main/Attention%2BResNet%20Results/attentionResNetResults.pdf}{Detailed Results}

\subsection{Analysis of BLEU-4 and Word Error Rate (WER)}

Typically, it is believed that a lower word error rate (WER) indicates superior accuracy in speech recognition. However, we observe the following:
\begin{itemize}
    \item The Fair prediction has a lower WER than the Good prediction.
    \item This demonstrates that true understanding of language depends on more than just:
    \begin{itemize}
        \item Low WER
        \item High accuracy
    \end{itemize}
\end{itemize}

On the contrary, BLEU-4 scores align better with human judgment because they consider:
\begin{enumerate}
    \item Adequacy – whether the full meaning of the source is conveyed.
    \item Fidelity – how accurately the translation matches the original.
    \item Fluency – how grammatically well-formed the sentence is.
\end{enumerate}
These three factors determine what makes a good sentence according to human judgment, which explains why BLEU-4 scores correlate well with human evaluation.

\subsection{Comparison: Vanilla RNN vs. Attention Model}

It can be observed that:
\begin{itemize}
    \item The attention model performs better than the vanilla RNN model on all automatic NLP metrics except METEOR.
    \item Human judgment also rates the attention model higher than the vanilla model.
\end{itemize}

Furthermore:
\begin{itemize}
    \item Due to its nature, the attention model is more descriptive than the vanilla model.
    \item The attention model enables better interpretability by showing:
    \begin{itemize}
        \item Various regions of the image.
        \item How different image regions contribute to specific words in the caption.
    \end{itemize}
    \item This interpretability makes it a strong candidate for tasks requiring explainability in AI.
\end{itemize}

\begin{figure}[htbp]
    \centering
    \includegraphics[width=0.92\textwidth]{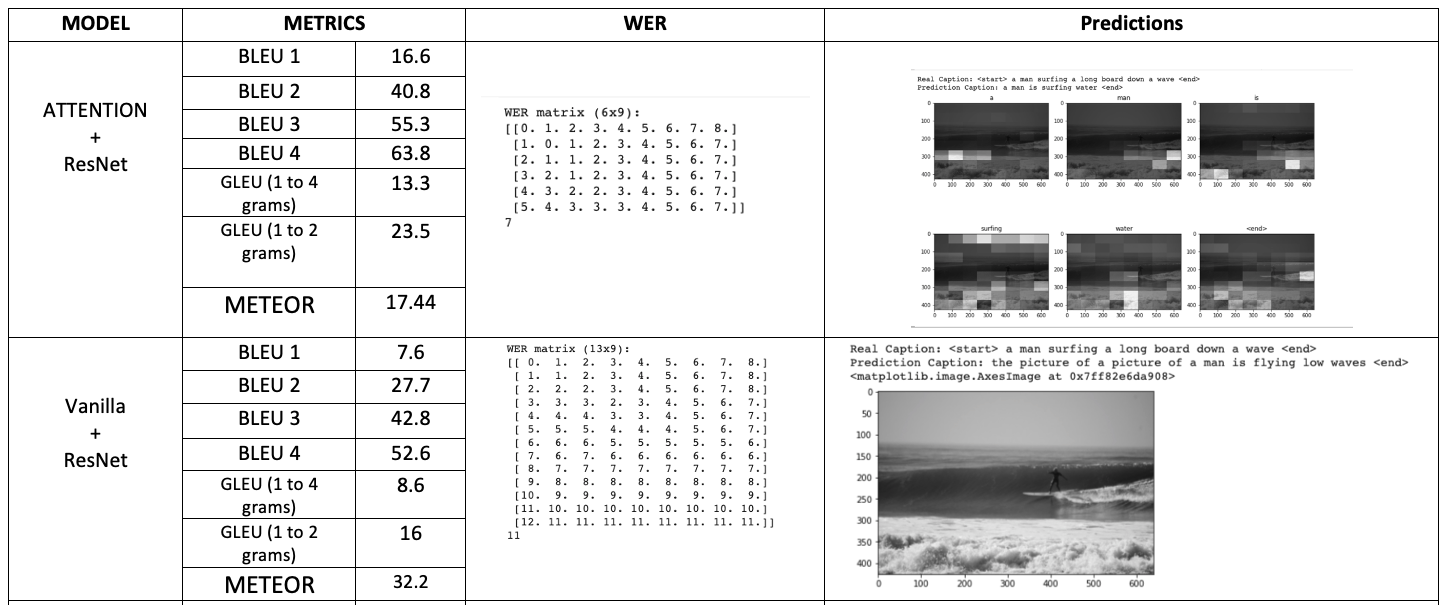}
    \caption{Comparison of Vanilla RNN vs. Attention-Based Model}
    \label{fig:compare}
\end{figure}

\noindent\textbf{View:} \href{https://github.com/HemanthTejaY/Deep-Learning-Image-Captioning---A-comparitive-study/blob/main/Images/detailed-res.pdf}{Detailed Results}

\section{Repository}

\noindent\textbf{GitHub Repository:} \href{https://github.com/HemanthTejaY/Deep-Learning-Image-Captioning---A-comparative-study}{Open Repository}

The complete source code, dataset preprocessing scripts, model architectures, and training details for this study are available in our GitHub repository. This repository provides a structured implementation of both Vanilla RNN-based and Attention-based image captioning models, along with scripts for evaluation metrics such as BLEU, METEOR, and CIDEr. Additionally, sample outputs, visualization scripts for attention heatmaps, and detailed documentation are included to facilitate reproducibility.

Researchers and developers interested in extending this work or applying it to other datasets can explore the repository for pre-trained models and training logs. Contributions, suggestions, and discussions are welcome through GitHub issues and pull requests.

\section{Conclusions}

Having implemented and studied the performance of both Vanilla RNN and Attention-based Image Captioning techniques on the MS COCO dataset, Natural Images, and Abstract Images, we can conclude that, in terms of performance on both human judgment and automatic metrics, the Attention Mechanism is superior to the Vanilla RNN Image Captioning on the COCO dataset and Natural Images with fewer subjects. However, both models perform similarly on abstract paintings and Natural Images with a large number of subjects.

Design considerations, such as the choice of CNN-Encoder, also affect the performance of the model. Moreover, we have observed that the currently used automatic metrics for image captioning judge caption quality by determining similarity between candidate and reference captions. As a result, these metrics fail to achieve adequate levels of correlation with human judgments at the sentence level, reflecting the fact that they do not fully capture the set of criteria humans use when evaluating caption quality.

We have accomplished our goal of comparing and understanding the architecture and performance aspects of deep learning-based image captioning models. In the process, we have learned about the interplay between Computer Vision and NLP, encoder-decoder architectures, loss functions in complex architectures, and the crucial ability to comprehend academic research papers in deep learning.


\begin{thebibliography}{9}

\bibitem{shetty2017}
Shetty, Rakshith, et al. “Speaking the Same Language: Matching Machine to Human Captions by Adversarial Training.” \textit{arXiv preprint arXiv:1703.10476}, Nov. 2017. \url{http://arxiv.org/abs/1703.10476}.

\bibitem{hossain2018}
Hossain, Md Zakir, et al. “A Comprehensive Survey of Deep Learning for Image Captioning.” \textit{arXiv preprint arXiv:1810.04020}, Oct. 2018. \url{http://arxiv.org/abs/1810.04020}.

\bibitem{tavakoli2017}
Tavakoli, Hamed R., et al. “Paying Attention to Descriptions Generated by Image Captioning Models.” \textit{arXiv preprint arXiv:1704.07434}, Aug. 2017. \url{http://arxiv.org/abs/1704.07434}.

\bibitem{vinyals2014}
Vinyals, Oriol, Alexander Toshev, Samy Bengio, and Dumitru Erhan. “Show and Tell: A Neural Image Caption Generator.” \textit{CoRR}, abs/1411.4555, 2014.

\bibitem{xu2015}
Xu, Kelvin, Jimmy Ba, Ryan Kiros, Kyunghyun Cho, Aaron C. Courville, Ruslan Salakhutdinov, Richard S. Zemel, and Yoshua Bengio. “Show, Attend and Tell: Neural Image Caption Generation with Visual Attention.” \textit{CoRR}, abs/1502.03044, 2015.

\bibitem{lin2014}
Lin, Tsung-Yi, Michael Maire, Serge Belongie, James Hays, Pietro Perona, Deva Ramanan, Piotr Dollar, and C. Lawrence Zitnick. “Microsoft COCO: Common Objects in Context.” In David Fleet, Tomas Pajdla, Bernt Schiele, and Tinne Tuytelaars, eds., \textit{Computer Vision – ECCV 2014}, pp. 740–755, Cham, 2014. Springer International Publishing.

\bibitem{kuznetsova2012}
Kuznetsova, Polina, Vicente Ordonez, Alexander C. Berg, Tamara L. Berg, and Yejin Choi. “Collective Generation of Natural Image Descriptions.” In \textit{Proceedings of the 2012 Conference}, pp. 359–368, 2012.

\bibitem{karpathy2017}
Karpathy, Andrej, and Li Fei-Fei. “Deep Visual-Semantic Alignments for Generating Image Descriptions.” \textit{IEEE Transactions on Pattern Analysis and Machine Intelligence}, vol. 39, no. 4, pp. 664–676, Apr. 2017.

\bibitem{pytorch}
PyTorch. “Loss Functions.” \url{https://ml-cheatsheet.readthedocs.io/en/latest/loss_functions.html}.

\end{thebibliography}
\end{document}